\pdfoutput=1

\documentclass[11pt]{article}

\usepackage{acl}

\usepackage{times}
\usepackage{latexsym}

\usepackage[T1]{fontenc}

\usepackage[utf8]{inputenc}

\usepackage{microtype}

\usepackage{inconsolata}

\usepackage{graphicx}
\usepackage{xcolor}
\usepackage{tcolorbox}

\newtcbox{\highlight}[1][red]{on line,
arc=4pt,colback=#1!20,
before upper={\rule[-3pt]{0pt}{10pt}},boxrule=0pt,
boxsep=0pt,leftrule=0pt,rightrule=0pt,toprule=0pt,bottomrule=0pt,left=3pt,right=3pt,top=2pt,bottom=1pt, colupper=black,fontupper=\ttfamily}

\definecolor{Document}{RGB}{117, 104, 236}
\definecolor{Group}{RGB}{225, 43, 107}
\definecolor{Topic}{RGB}{251, 132, 57}
\definecolor{Word}{RGB}{49, 170, 190}

\title{\includegraphics[scale=0.2,trim=0 2.7cm 0 0]{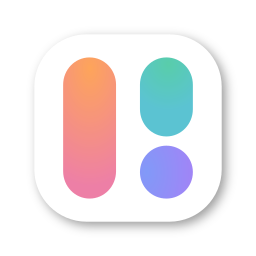} \texttt{topicwizard} - a Modern, Model-agnostic Framework for Topic Model Visualization and Interpretation}

\author{
  Márton Kardos \\
  Aarhus University\\ 
  \texttt{\small martonkardos@cas.au.dk} \\\And
  Kenneth C. Enevoldsen \\
  Aarhus University\\ 
  \texttt{\small kenneth.enevoldsen@cas.au.dk} \\\And
  Kristoffer Laigaard Nielbo \\
  Aarhus University \\
  \texttt{\small kln@cas.au.dk} 
  }

\begin{document}
\maketitle
\begin{abstract}
Topic models are statistical tools that allow their users to gain qualitative and quantitative insights into the contents of textual corpora without the need for close reading \citep{quantitative_text_analysis}.
They can be applied in a wide range of settings from discourse analysis \citep{discourse_analysis}, through pretraining data curation \citep{pretraining_data_curation}, to text filtering \citep{filtering}.
Topic models are typically parameter-rich, complex models, and interpreting these parameters can be challenging for their users.
It is typical practice for users to interpret topics based on the top 10 highest ranking terms on a given topic.
This \textit{list-of-words} approach, however, gives users a limited and biased picture of the content of topics \citep{interpretation_for_scholarly_analysis}.
Thoughtful user interface design and visualizations can help users gain a more complete and accurate understanding of topic models' output.
While some visualization utilities do exist for topic models,
these are typically limited to a certain type of topic model.
We introduce \texttt{topicwizard} \footnote{\url{https://github.com/x-tabdeveloping/topicwizard}},
a framework for model-agnostic topic model interpretation,
that provides intuitive and interactive tools that help users examine the complex semantic relations between documents,
words and topics learned by topic models.
\end{abstract}

\section{Introduction}

Topic models are statistical instruments, which have been developed to uncover human-interpretable topics in corpora of text \citep{probabilistic_topic_models}.
These methods have allowed analysts gain insights into the contents of large corpora,
the manual reading of which would be impractical or impossible.
Topic models also often offer a more impartial account of a corpus' content \citep{quantitative_text_analysis}.

Typically, topic models' outputs are presented to users in the form of the highest-ranking words and perhaps documents on a given topic.
While this allows users to gain a superficial understanding of a topic,
one might miss crucial details, and a lot of nuances, when topic models are exmined this way \citep{interpretation_for_scholarly_analysis}.
We suggest that topic models capture more detailed information about topics than simple word lists convey, and that carefully designed interfaces can help users better explore this complexity.

\subsection{Topic Models are Diverse}

While topic models all carry out a similar task,
they can also be very different from each other in how they conceptualize topic discovery.

Topic models originally relied on a bag-of-words model of documents where they are represented as sparse vectors of word-occurrence counts,
with an optionally applied weighting scheme, such as tf-idf.
Most commonly, these models either discover topics by matrix factorization (\citealt{nmf}, \citealt{lsa})
or by fitting a probabilistic generative model over these representations (\citealt{lda}, \citealt{gsdmm}, \citealt{plsi}) or biterms \citep{btm}.

\begin{figure}[htbp]
    \centering
    \includegraphics[width=\linewidth]{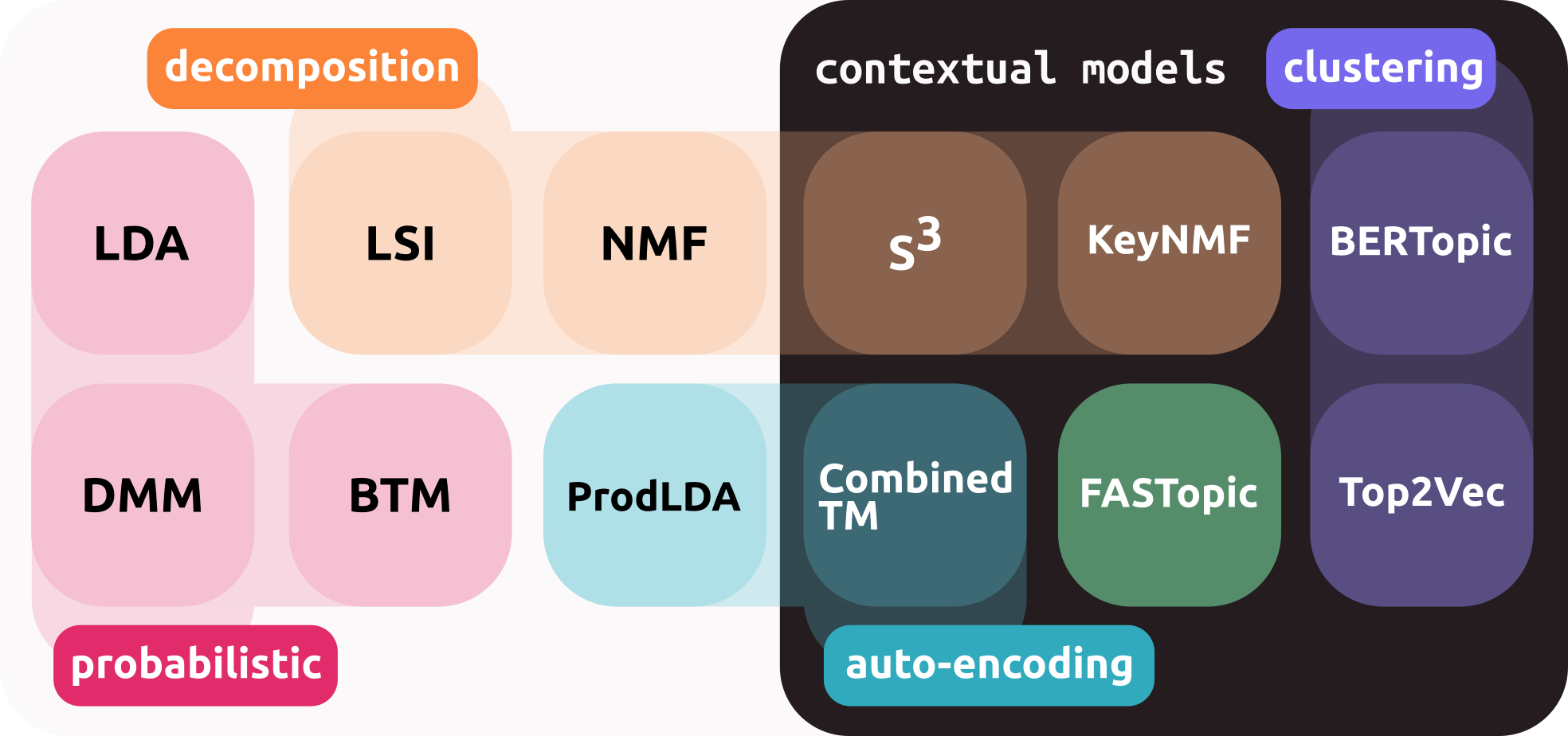}
    \caption{A Simplified Taxonomy of Topic Models
    }
    \label{fig:topic_taxonomy}
\end{figure}

More recent topic models, however, also rely on context-sensitive, dense text representations from neural networks \citep{sentence_transformers}.
These models can conceptualize topic discovery as document clustering and post-hoc term importance estimation (\citealt{bertopic}, \citealt{top2vec}), 
document generation with amortized variational inference (autoencoders) (\citealt{ctm}, \citealt{zeroshot_tm}),
semantic relation reconstruction \citep{fastopic},
or semantic decomposition (\citealt{s3}, \citealt{keynmf}).

\subsection{Topic Models are Alike}
Despite these differences, all topic models have a lot in common.
Topic models, in essence, learn a three-way relationship between \highlight[Word]{words}, \highlight[Document]{documents} and \highlight[Topic]{topics}.

\begin{figure}[htbp]
    \centering
    \includegraphics[width=\linewidth]{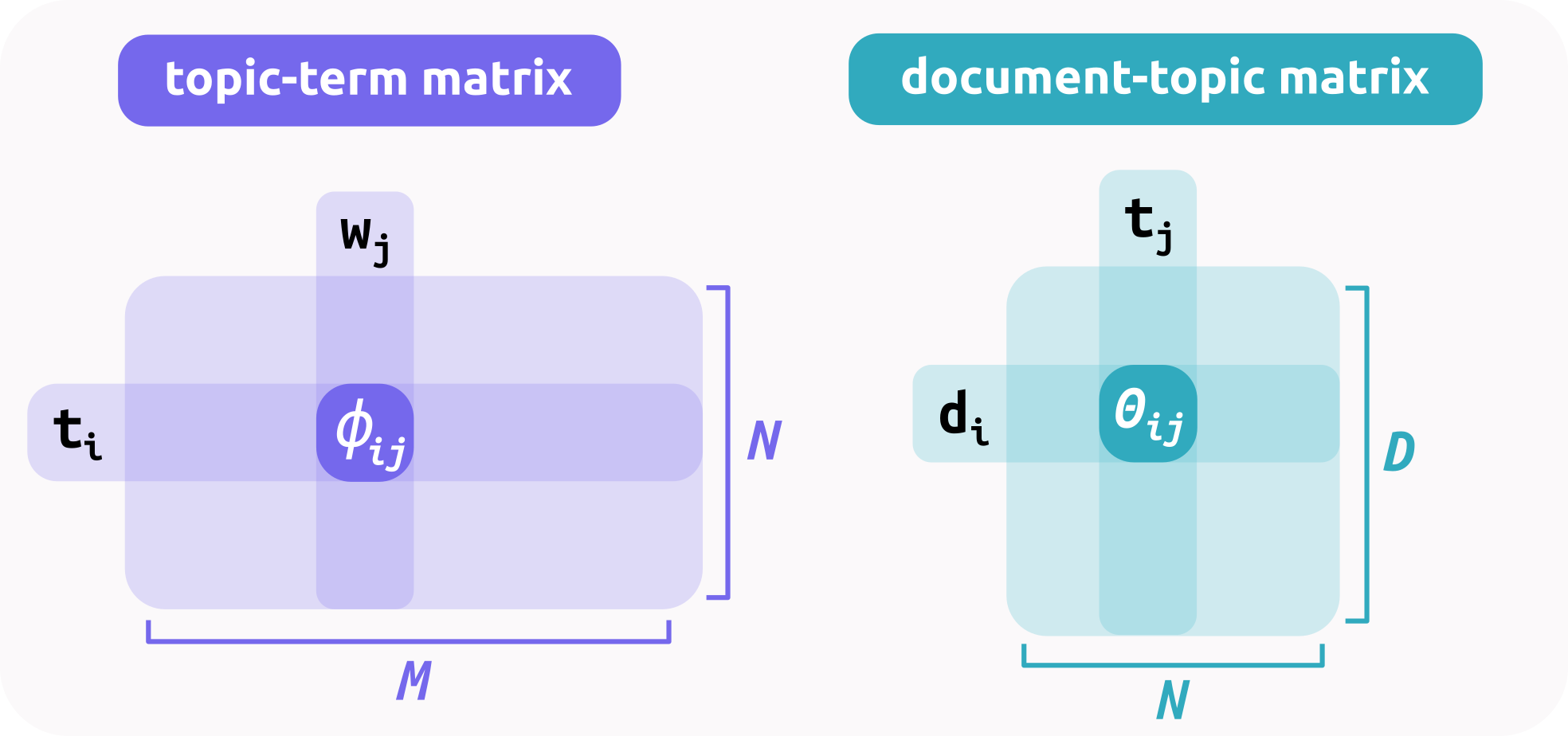}
    \caption{Common Components Computed by Topic Models}
    \label{fig:shared_components}
\end{figure}

All topic models have a method for extracting the $K$ most relevant words from the discovered topics.
These top $K$ words are calculated from a \textbf{topic-term matrix} ($\phi$),
which is either inferred as part of topic discovery.
This matrix has $N$ rows, corresponding to the number of topics,
and $M$ columns corresponding to the size of the model's vocabulary.
In addition, models compute a \textbf{document-topic-matrix} ($\Theta$), where rows represent the $D$ documents in the corpus,
while the $N$ columns represent topics.
This matrix contains the importance/relevance of a topic in a document.

\subsection{Contribution}

We introduce \texttt{topicwizard}, a model-agnostic topic model visualization framework that allows users to investigate complex semantic relations between words, documents and topics in their corpora.
\texttt{topicwizard} is natively compatible with topic modelling libraries, which use the scikit-learn API \citep{scikit-learn}, such as tweetopic \citep{tweetopic} and Turftopic \citep{turftopic} and comes with compatibility layers for Gensim and BERTopic.

\section{Related Work}

\begin{figure*}[ht]
    \includegraphics[width=\linewidth]{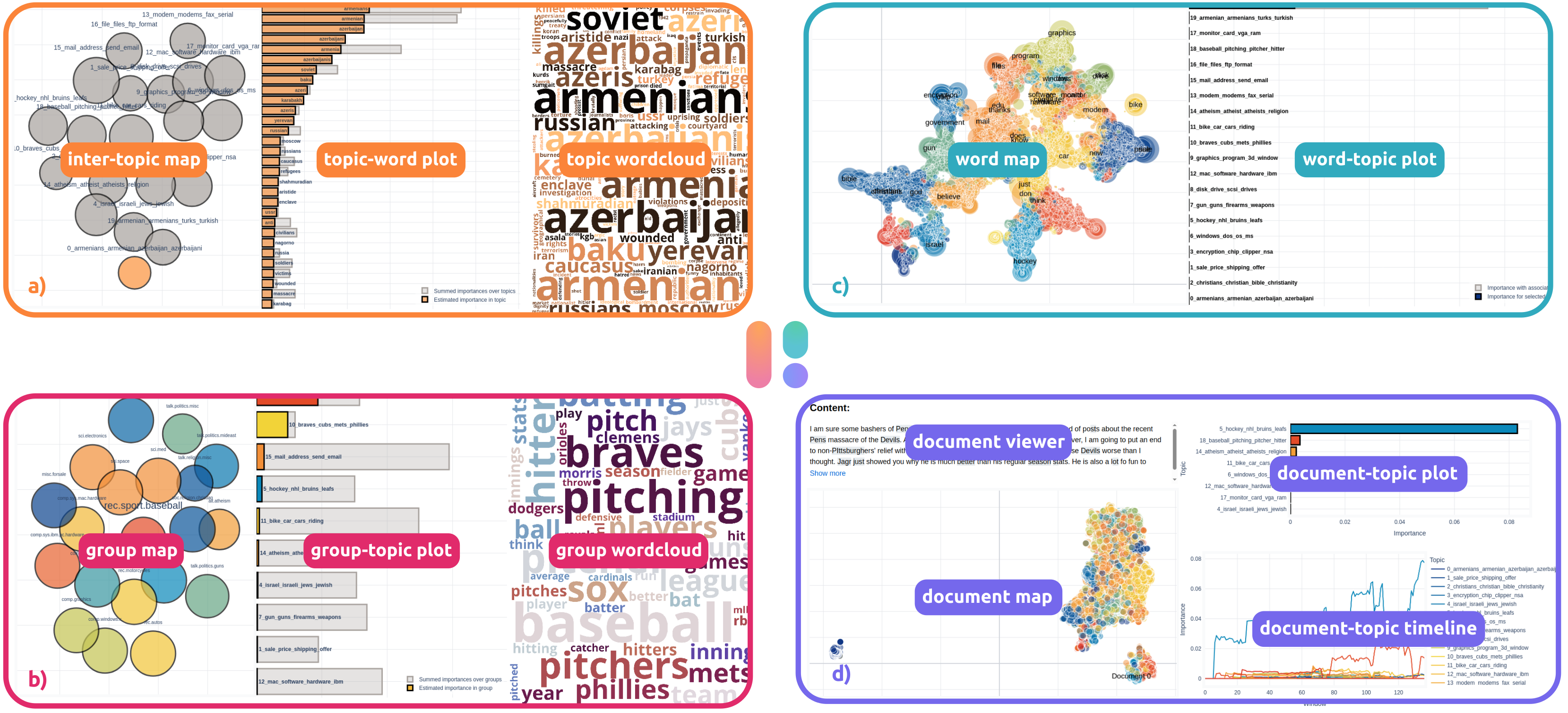}
    \caption{
        An overview of visualizations and pages in the \texttt{topicwizard} framework \\
        \textit{All visualizations were produced using KeyNMF \citep{keynmf}}
    }
    \label{fig:overview}
\end{figure*}

Due to Latent Dirichlet Allocation's (LDA) popularity, a considerable amount of work has been dedicated to visualizing and interpreting its outputs.
\citet{interpretation_and_trust} discuss best practices and design considerations for visualization and interpretation systems for LDA.
\citet{termite} introduced the Termite system for interactively visualizing and interpreting LDA output.
The main visualization in Termite is a stylized version of the topic-term matrix (see Figure \ref{fig:termite}),
where circles of different size are at the intersection of terms and topics indicating their importance.
The authors also propose a scheme for selecting the most topically salient words, since displaying all words in the corpus would not be feasible.
As a consequence, Termite can only display a limited number of words. 
Additionally, Termite is no longer under active maintenance \footnote{The Termite repository on Github was last committed to 11 years prior to the writing of this article}.

LDAvis \citep{ldavis} is an interactive visualization R package for LDA (see Figure \ref{fig:ldavis}).
LDAvis combines elements of previous topic visualization systems,
including an inter-topic distance map, term distribution plots,
and a term-weighting scheme to show only the most specific and (\textit{relevant}) terms.
Similar to Termite, the original LDAvis package is no longer maintained.
Its Python port, PyLDAvis, receives occasional updates, but does not enjoy feature parity with the original package.

Notable visualization utilities are also included in the BERTopic library \citep{bertopic},
which boasts model-specific plotting functions, such as an inter-topic map, document cluster visualizations, and term distribution bar-charts.
Similarly, Turftopic \citep{turftopic} also contains model-specific visualization utilites for a number of models,
including cluster maps,
concept compasses for $S^3$ \citep{s3} and interactive timeline plots for dynamic topic models.
While these visualizations are useful, they are typically of limited interactivity, and are limited to a particular type of model.

\section{\includegraphics[scale=0.1,trim=2cm 1.8cm 0cm 0cm]{figures/logo.png} \texttt{topicwizard}}

To address these challenges, we outline \texttt{topicwizard}, a novel system for topic model interpretation.
Our framework is model-agnostic, allows users to investigate topic models from a number of distinct perspectives, and is highly interactive,
thereby providing a more complete picture of topic models' output,

\subsection{Topic Models Learn Topic Representations}

Topic models' primary objective is to discover latent themes in a corpus.
Being able to understand what concepts make up such topics, and how these topics are related is perhaps the most important aspect of interpreting topic models.

In \texttt{topicwizard} (see Figure \ref{fig:overview}a), similar to \citet{ldavis} an
\highlight[Topic]{inter-topic map}
is displayed, which shows the relative distances of topics to each other.
While \citet{ldavis} utilize PCA for this visualization, 
projections in \texttt{topicwizard} are calculated with UMAP \citep{umap}, since it is better at capturing local structure.
Topic importance is calculated in the following manner:

$$s_t = \sum_d^D{\Theta_{dt} \cdot |d|}$$
where $\Theta_{dt}$ is the importance of topic $t$ and document $d$ and $|d|$ is the number of terms in a given document, and $D$ is the size of the corpus. 

This metric dictates the size of topics on the graph.

To provide users with insights about topics' word content, the \highlight[Topic]{topic-word plot} displays the distribution of the highest ranking words for a given topic, and also how globally prevelant these words are across topics
\footnote{Unlike LDAvis, we do not compute \textit{relevance} scores, since they rely on the assumption that $\phi$ contains word probabilities.}.
Since 10-20 words are rarely enough to give a complete picture of the words relevant to a topic, a more comprehensive \highlight[Topic]{topic wordcloud} is also displayed
To aid further analysis, users can also manually name topics on this page.

\subsection{Topic Models Learn Word Embeddings}

While topic models' are mainly oriented at discovering topics, they also implicitly learn meaningful representation of words within the corpus.
Each column of the topic-term matrix can technically be thought of as a semantic embedding for a given word, with the dimensions being interpretable.
This implicit learning of word representations allows us to examine words' relation to each other in a corpus, without explicit reference to the topics.

In \texttt{topicwizard} (see Figure \ref{fig:overview}c), a \highlight[Word]{word map} is displayed to users,
allowing them to quickly and interactively investigate the semantic landscape of words in their corpus.
Word positions are calculated by projecting word embeddings to two dimensions using UMAP.

Word embeddings are useful for investigating associative relations in corpora, and have been used for a variety purposes such as query expansion \citep{query_expansion},
or to uncover authorship patterns in literature \citep{grundtvig_og_monstrene}.
Clicking on a word on the word map highlights the words most closely related to the selected one and displays the topical distribution of the selected term and its neighbourhood on the \highlight[Word]{word-topic plot}.
Displaying closely associated words with the selected keywords in topic models can give practitioners a more nuanced picture of word use \citep{appeal_to_political_sentiment}.

\subsection{Topic Models Organize Documents}

An important aspect of topic models is that they learn a representation of documents in the corpus they are fitted on.
Document representations discovered by topic models were historically used for a number of purposes,
including retrieval \citep{retrieval}, and studying information dynamics \citep{infodynamics}.

In \texttt{topicwizard} (see Figure \ref{fig:overview}d), a \highlight[Document]{document map} is displayed, where document's UMAP-projected embeddings can be seen,
and documents are coloured based on most prevalent topic.
In the case of BoW models, these representations are derived from the document-topic matrix, while with contextual models, the pre-computed sentence embeddings are used.

Secondly, individual documents' contents can be investigated on a \highlight[Document]{document-topic plot}, which displays the distribution of the most relevant topics,
a \highlight[Document]{document-topic timeline}, which displays how the topical content changes throughout the course of the document and a 
\highlight[Document]{document viewer}, where a snippet of the document is displayed, and the most topically relevant words are highighted.
The combination of these document inspection utilities can help users ground and verify topic models' output in the documents themselves,
which elevates trust \citep{interpretation_and_trust}.
Additionally, this interface encourages close reading, which provides additional insight into the corpus' content.

\subsection{Topics Augment User-Defined Groups}

Commonly, users of topic models also have some externally defined grouping of documents, which might be relevant for their analyses.
This could be binning documents by time period, predefined categories or place of origin.
While most topic models do not utilize external labels, meaningful inferences can be made about topics' relation to these labels post-hoc.

An important part of this process is to compute a \textbf{group-topic matrix},
the cells of which contain the summed importance of a given topic for documents in a given group:

$$G_{ij} =\sum_k^D{\Theta_{kj} \cdot I(g_k = i)} $$ where
$G_{ij}$ is the importance of group $i$ for topic $j$, $g_k$ is the group label of document $k$, and $I(g_k=i)$ is the indicator function.

In \texttt{topicwizard} (see Figure \ref{fig:overview}b), semantic distances between user-defined groups can be seen on the \highlight[Group]{group map}, where group-topic representations are projected to 2D space using UMAP.
Groups are coloured based on the dominant topic in the group.
Topic distributions in groups can be seen on the \highlight[Group]{group-topic plot},
and groups' lexical content can be examined in detail on the \highlight[Group]{group wordcloud} to the right.

\subsection{Software Design Considerations}

The \texttt{topicwizard} Python package was designed with both research and enterprise use in mind.
As such, our goal was to develop a package that is accessible to new users and sufficiently flexible to accommodate specific use cases -- ranging from academic writing and technical reporting to enabeling business analysts to interact with topic models via a web interface.

The \textbf{Web Application} (see Figures \ref{fig:topicwizard_web_app_topics} - \ref{fig:topicwizard_web_app_groups}) was designed to make topic model interpretation as seemless and quick as possible, in as many environments as possible, including Jupyter notebooks, in the browser, or deployed to the cloud.
which produces a readily deployable Docker project to a specified folder.

The \textbf{Figures API} makes it trivial for our users to produce specific figures tailored to their needs.
This is especially crucial for producing publications, since some colour schemes, fonts or aspect ratios, while appropriate for an interactive web application, might not be visually appealing in a static document.

\section{Conclusion}

This paper introduces \texttt{topicwizard}, a comprehensive, interactive, and model-agnostic topic model visualization framework.
Our framework is a notable extension over previous topic model visualization systems,
thanks to a) supporting a much wider range of models 
b) allowing users to ground topic models in the corpus, and investigate them from numerous angles and
c) being flexible, actively supported, and production-ready.
The \texttt{topicwizard} software package has so far been downloaded more than 45000 times from PyPI,
demonstrating that practitioners have already found it useful.

\newpage

\section*{Limitations}

While \texttt{topicwizard} is the most comprehensive topic model visualization tool to date, it still lacks coverage of a number of aspects of topic modelling.
It, for instance, does not have visualization utilities for dynamic, hierarchical and supervised topic models.
This is a clear limitation and will have to be addressed in future package releases.

Our framework, as of now, does not provide any utilities for comparing outputs from different topic models either.
This is yet another aspect that future work should address.

Furthermore, while we consider model-angosticity to be one of the strengths of our approach, it does, to an extent, limit its usefulness for certain models.
Certain visualizations, such as concept compasses, might be highly useful tools for examining the output of Semantic Signal Separation, but their utility might be limited for clustering topic models.
We encourage our users, therefore, to use \texttt{topicwizard} in tandem with model-specific interpretation utilities from libraries such as BERTopic or Turftopic.

\bibliography{anthology, custom}

\appendix

\section{Screenshots}

This section contains screenshots of both \texttt{topicwizard} and systems that preceeded it.
\label{sec:screenshots}

\begin{figure*}[h]
    \includegraphics[width=\linewidth]{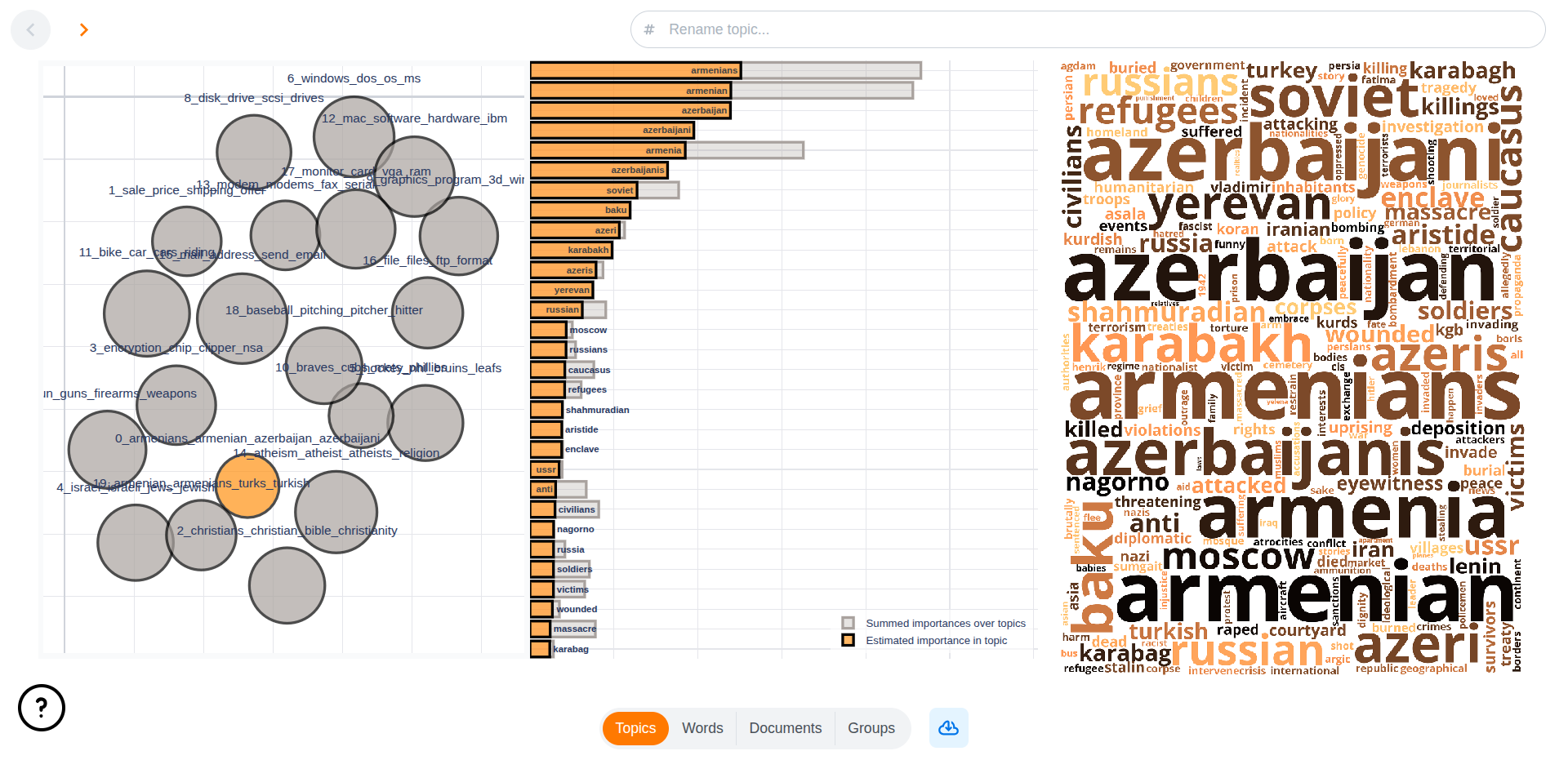}
    \caption{Screenshot of the \highlight[Topic]{Topics} page in the \texttt{topicwizard} Web Application
    }
    \label{fig:topicwizard_web_app_topics}
\end{figure*}

\begin{figure*}[h]
    \includegraphics[width=\linewidth]{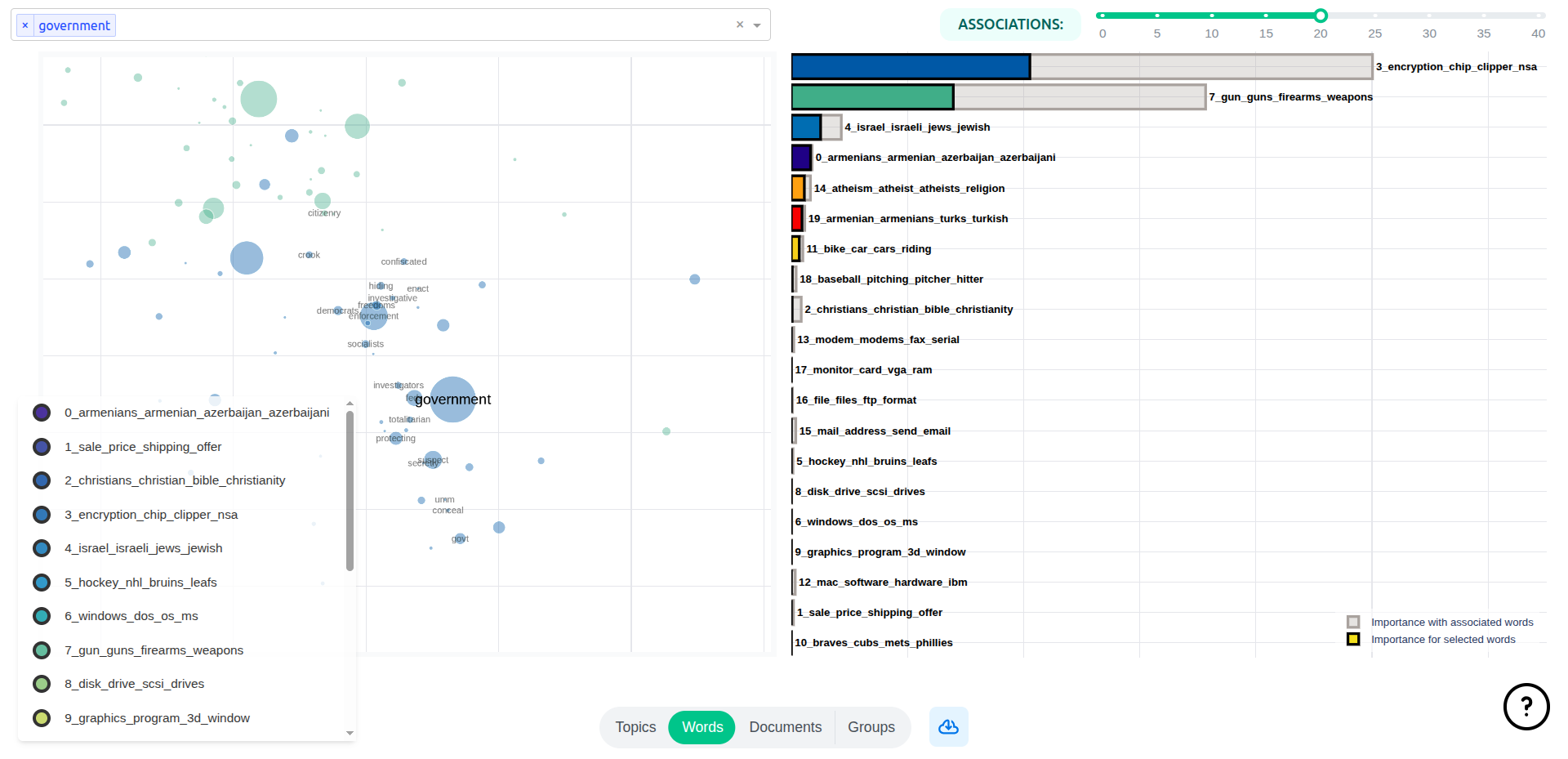}
    \caption{Screenshot of the \highlight[Word]{Words} page in the \texttt{topicwizard} Web Application
    }
    \label{fig:topicwizard_web_app_words}
\end{figure*}

\begin{figure*}[h]
    \includegraphics[width=\linewidth]{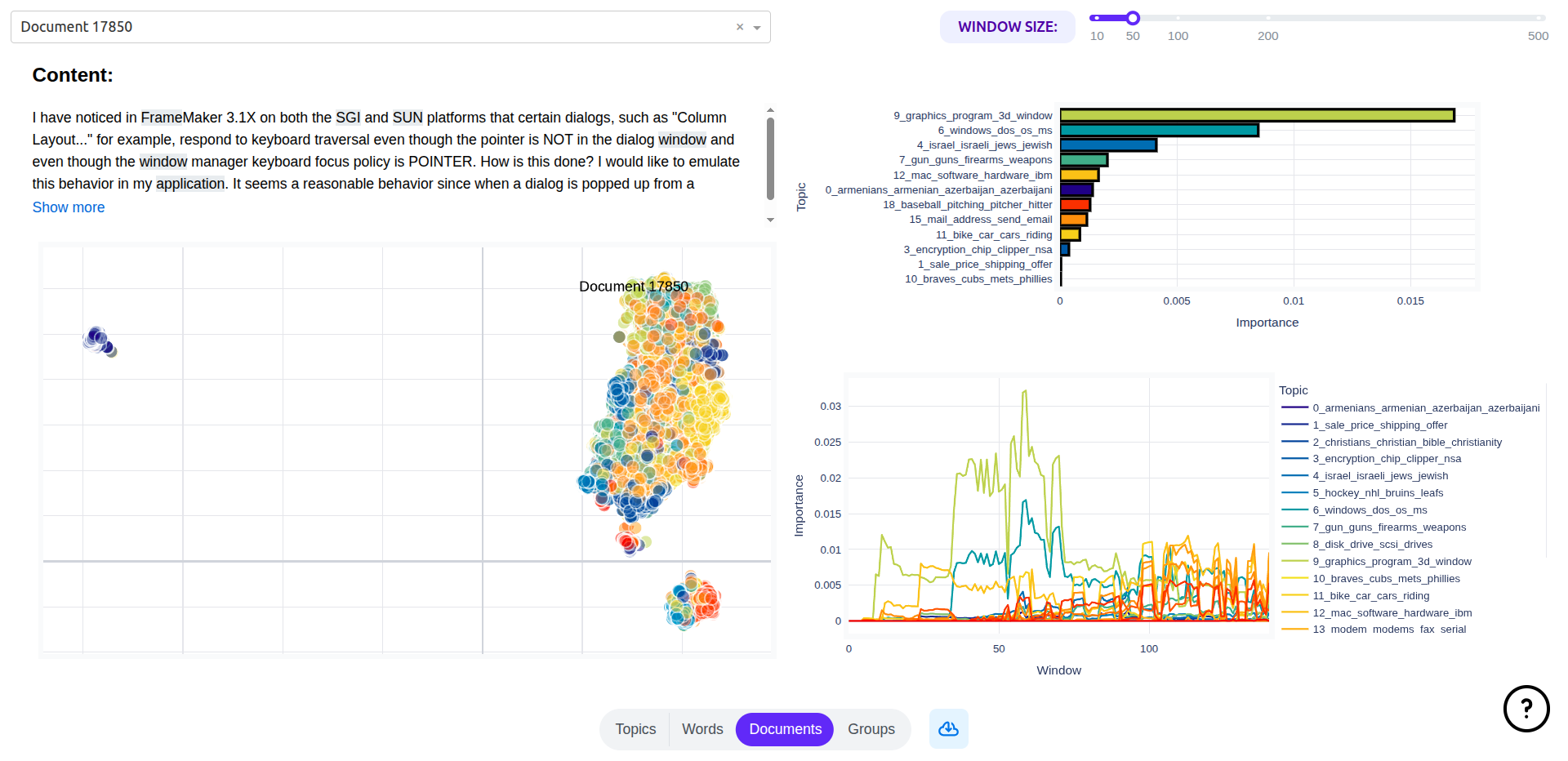}
    \caption{Screenshot of the \highlight[Document]{Documents} page in the \texttt{topicwizard} Web Application
    }
    \label{fig:topicwizard_web_app_documents}
\end{figure*}

\begin{figure*}[h]
    \includegraphics[width=\linewidth]{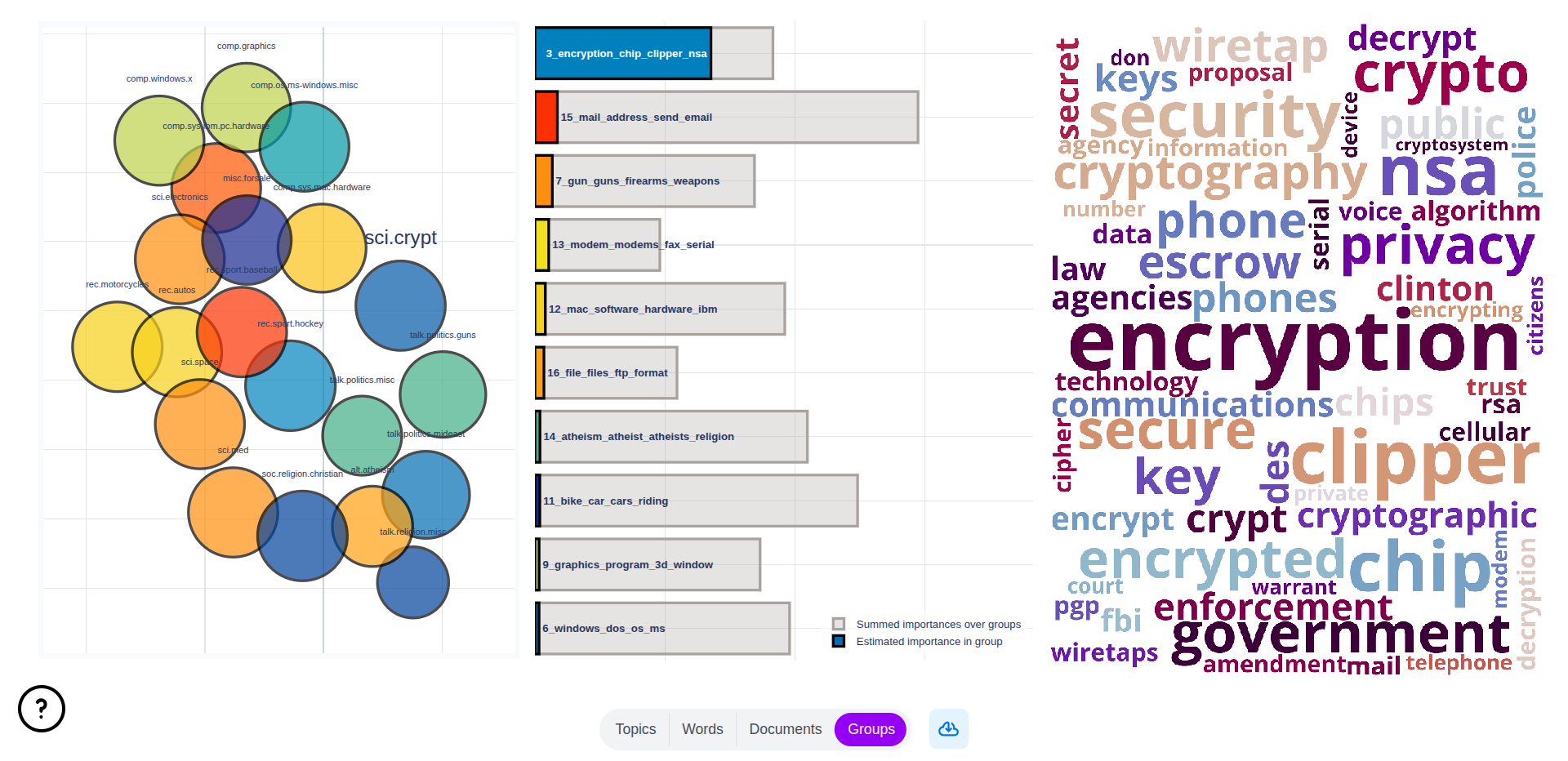}
    \caption{Screenshot of the \highlight[Group]{Groups} page in the \texttt{topicwizard} Web Application
    }
    \label{fig:topicwizard_web_app_groups}
\end{figure*}

\begin{figure*}[h]
    \includegraphics[width=\linewidth]{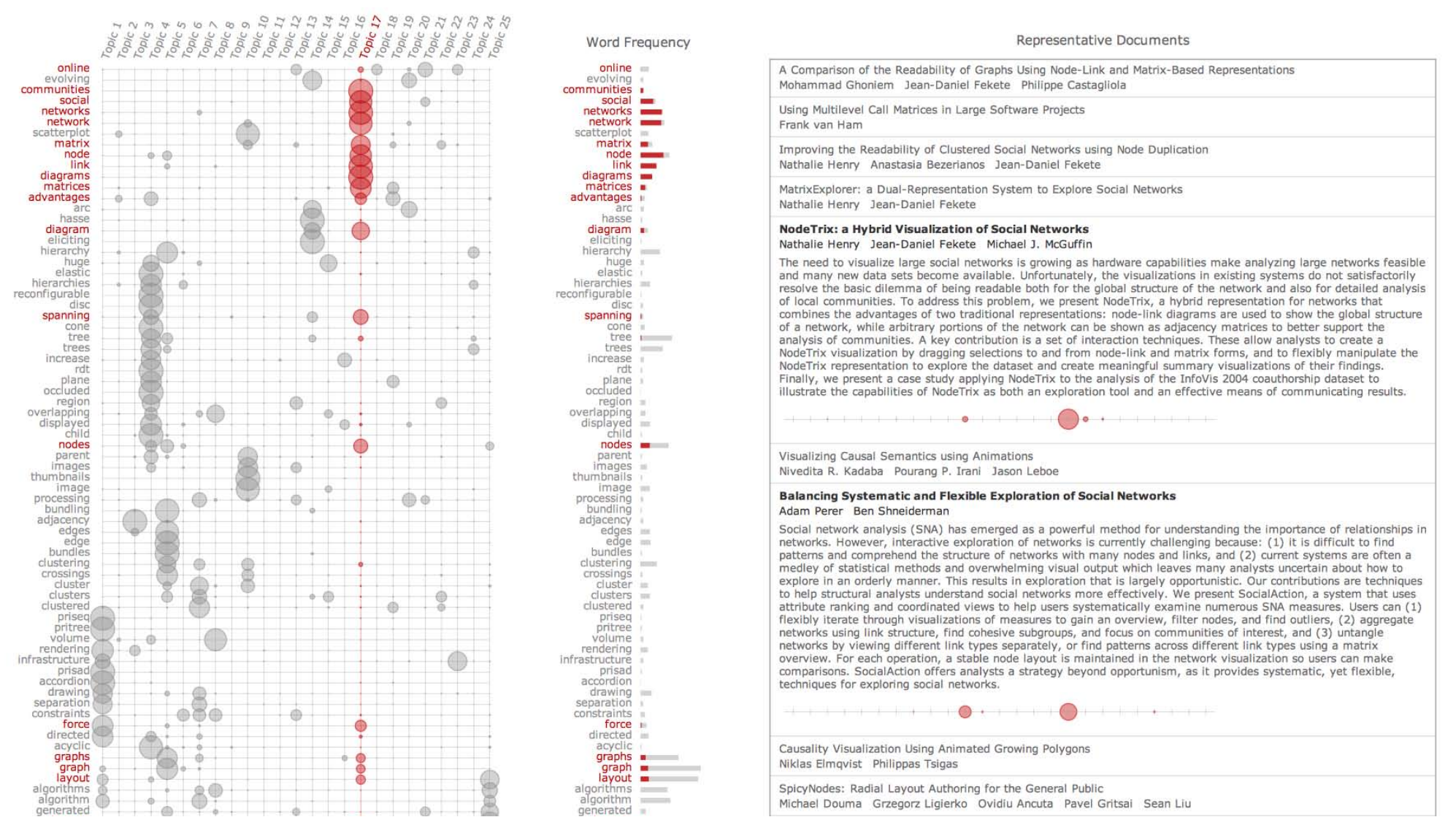}
    \caption{Screenshot of the Termite System\\
        Figure from \citep{termite}
    }
    \label{fig:termite}
\end{figure*}

\begin{figure*}[h]
    \includegraphics[width=\linewidth]{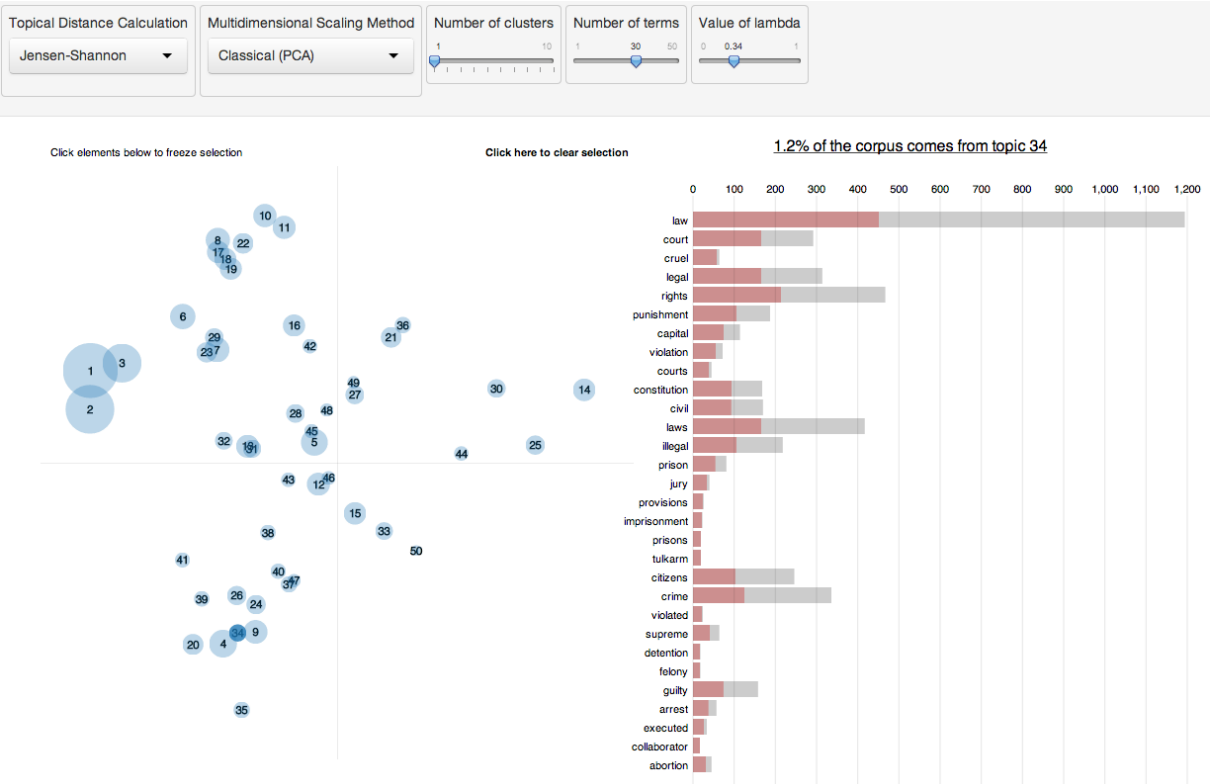}
    \caption{Screenshot of LDAvis\\
        Figure from \citep{ldavis}
    }
    \label{fig:ldavis}
\end{figure*}

\end{document}